\documentclass[10pt,journal,letterpaper,compsoc]{IEEEtran}

\usepackage{graphicx}
\usepackage{dcolumn}
\usepackage{bm}

\begin{document}


\title{Statistical analysis of the Indus script using $n$-grams}

\author{Nisha Yadav,~\IEEEmembership{}
        Hrishikesh Joglekar,~\IEEEmembership{}
        Rajesh P. N. Rao,~\IEEEmembership{}
        M. N. Vahia,~\IEEEmembership{} 
        Iravatham Mahadevan~\IEEEmembership{}
        and~R.~Adhikari~\IEEEmembership{}

\IEEEcompsocitemizethanks{\IEEEcompsocthanksitem Nisha Yadav and Mayank Vahia are with the Tata Institute of Fundamental Research Homi Bhabha Road, Colaba, Mumbai 400 005, India and Centre for Excellence in Basic Sciences, Mumbai, India
E-mail: y$\_$nisha@tifr.res.in 
\IEEEcompsocthanksitem Hrishikesh Joglekar is with Oracle, Hyderabad 500 081, India
\IEEEcompsocthanksitem Rajesh Rao is with Department of Computer Science \& Engineering,  University of Washington, Seattle, WA 98195, USA 
\IEEEcompsocthanksitem Iravatham Mahadevan is with Indus Research Centre,  Roja Muthiah Research Library, Chennai 600 113, India  
\IEEEcompsocthanksitem R. Adhikari (corresponding author) is with the Institute of  Mathematical Sciences, Chennai 600 113, India 
E-mail: rjoy@imsc.res.in 
}
\thanks{Manuscript received December, 2008}}

\markboth{Statistical Analysis of Indus Script}%
{Shell \MakeLowercase{Yadav \textit{et al.}}: Indus Script}
\IEEEcompsoctitleabstractindextext{%

\begin{abstract}

The Indus script is one of the major undeciphered scripts of the ancient world. The small size of the corpus, the absence of bilingual texts, and the lack of definite knowledge of the underlying language has frustrated efforts at decipherment since the discovery of the remains of the Indus civilisation. Recently, some researchers have questioned the premise that the Indus script encodes spoken language. Building on previous statistical approaches, we apply the tools of statistical language processing, specifically $n$-gram Markov chains, to analyse the Indus script for syntax. Our main results are that the script has well-defined signs which begin and end texts, that there is directionality and strong correlations in the sign order, and that there are groups of signs which appear to have identical syntactic function. All these require no {\it a priori} suppositions regarding the syntactic or semantic content of the signs, but follow directly from the statistical analysis. Using information theoretic measures, we find the information in the script to be intermediate between that of a completely random and a completely fixed ordering of signs. Our study reveals that the Indus script is a structured sign system showing features of a formal language, but, at present, cannot conclusively establish that it encodes {\it natural} language. Our $n$-gram Markov model is useful for predicting signs which are missing or illegible in a corpus of Indus texts. This work forms the basis for the development of a stochastic grammar which can be used to explore the syntax of the Indus script in greater detail.

\end{abstract}

}
\maketitle

\IEEEdisplaynotcompsoctitleabstractindextext

\section{Introduction}
The civilisation that flourished from 7000 BCE to 1500 BCE  in the valley of the river Indus and its surroundings was the first urban civilisation in the Indian subcontinent. At its peak, in the period between 2600 BCE and 1900 BCE, it covered approximately a million square kilometers \cite{Possehl2002}, making it the largest urban civilisation of the ancient world. The remains of the civilisation were first found in Harappa and, following historical convention, is called the Harappan civilisation. The terms Harappan civilisation and Indus civilisation are used interchangeably in this paper.

\begin{figure}[h]
\center
	\includegraphics[width=6cm]{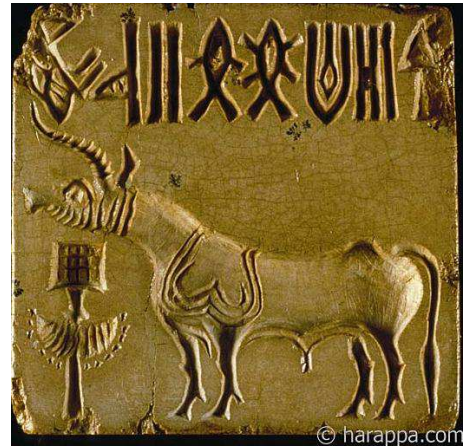}
	\caption{\label{fig:unicornseal}An example of an Indus seal, showing the three typical components: the Indus script at the top, a field symbol in the middle, and a decorated object at the bottom left. Here, since the script is embossed on a seal, it is to be read from the left to the right, whereas on the sealing, which are impressions of the seal, it is read from the right to the left. For the most part, the seals are typically between $1$ to $2$ square inches in size.}
\end{figure}

The Indus people used a system of signs, which has mainly survived on seals (see Fig.~\ref{fig:unicornseal} for an example), pottery, and other artifacts made of durable materials such as stone, terracotta, copper. The sign system is yet to be deciphered. The use of these signs is usually in short texts, numbering not more than $14$ signs in a single line of text. Around $400$ distinct signs have been identified \cite{M77,Parpola1994}, though some researchers identify up to $676$ distinct signs \cite{Bryan}. The total texts number to about $3000$. The lack of decipherment of the sign system is attributed to the paucity of the working material, the absence of any bilingual texts that might aid in decipherment, and a lack of definite knowledge of the underlying language, assuming the script was used to write language. The sign system of the Indus remains controversial, with contested claims of decipherment, and even the suggestion that it does not encode language \cite{Farmer}. Here we shall use the terms sign system and script interchangeably to refer to the Indus script.

The main methodological difficulty in attempting any interpretation of the script is that, due to the paucity of information on the context of the writing, one is perforce required to make an {\it a priori} assumption regarding the information the script is meant to convey. Thus, the range of opinion on what the script encodes varies from an Indo-Aryan language \cite{Rao},  a Dravidian language \cite{Parpola1994}, a purely numerical system \cite{Subbarayappa}, to non-linguistic symbols \cite{Farmer}. There is no consensus on any of the above interpretations. 

A line of research, which does not require such {\it a priori} assumptions, is the statistical approach. At its heart lies the notion of recognising patterns through counting. While such an approach cannot shed light on the semantics of the script, it can reveal important features of its syntax. Research using the statistical approach was initiated by a Soviet team in 1965, further developed by a Finnish team in 1969 (for review of various attempts see \cite{Possehl1996, Mahadevan2002, Parpola2005}), continued by Siromoney \cite{Gift} in the 1980s and followed up more recently by some of us \cite{Yadav1, Yadav2, Rajesh}. 

In this article, we bring to bear the tools of $n$-gram modelling for a thorough statistical analysis of sequences in the Indus script. An $n$-gram is a sequence of $n$ units, or tokens. $N$-gram models provide a very useful method of modelling sequences\addtocounter{footnote}{0}\footnote{The examples of English language used in the paper to explain various concepts related to n-gram modelling are given for the ease of comprehension, as a representative of the general class of sequences. These linguistic examples do not imply that any linguistic assumption has been made for the Indus script.}, whether they be the sequence of letters or words in a natural language, the base pairs in the genetic code, or the notes in a musical score. This generality is possible because $n$-gram models are indifferent to the semantic content of the units or tokens making up the sequence (the words, the letters, the base pairs or the notes) but, nonetheless, reveal the syntax, if any, that the sequences follow. The $n$-gram approach, then, forms the ideal framework in which the Indus script can be analysed without making any {\it a priori} assumptions. 

Our work in the paper uses a bigram model, that is an $n$-gram model with $n=2$, to model the sequence of signs in the Indus script. Our analysis reveals that the sign system of the Indus is structured, with specific text beginner and text ender signs, that there is directionality, strong correlations in the order of the signs, and groups of signs which appear to have the same syntactic function. In Section~\ref{sec:corpus}, we provide details of the corpus used in this study and present results of an empirical statistical analysis of the corpus. The empirical analysis shows that a small number of signs account for the majority of the usage, a fact we are able to capture by fitting the frequency distribution of signs to the well-known Zipf-Mandelbrot law. We note in passing, that word frequency distributions in natural languages also follow this law. In Section~\ref{sec:ngrams} we provide a summary of the theory of $n$-gram models,  paying special attention to the problem of unseen $n$-grams. We describe how this problem is resolved using smoothing and backoff algorithms and demonstrate how this modifies the statistics obtained from the empirical analysis. Our main results, summarised above,  are presented in Section~\ref{sec:bigram} using a bigram model. The Indus script corpora contains many instances of texts in which signs are damaged beyond recognition or otherwise illegible. The bigram model can be used to suggest a set of signs, ranked by probability,  for the restoration of the damaged or illegible signs. In Section~\ref{sec:filling} we outline the theory behind this method and provide examples on known texts from the corpus. In the final section, we discuss future work based on the $n$-gram approach and end with a summary of the present work.

\section{The corpus and empirical analysis}
\label{sec:corpus}

Any work on the Indus script,  statistical or otherwise, must begin with a corpus. Three major corpora of the Indus texts, by Mahadevan \cite{M77}, Parpola \cite{Parpola1994} and Wells \cite{Bryan} are available. We use the electronic concordance of Mahadevan, henceforth referred to as M77, which records $417$ unique signs\addtocounter{footnote}{0}\footnote{The serial number of the signs used in this paper is as given by Mahadevan in his concordance (M77). As a convention followed in the present paper, the texts depicted by pictures are to be read from right to left, whereas the texts represented by just strings of sign numbers are to be read from left to right (see M77 for discussion on direction of texts).} in $3573$ lines of $2906$ texts. In order to remove ambiguity, an Extended Basic Unique Data Set (EBUDS) is created by removing from the concordance all texts containing lost, damaged, illegible, or doubtfully read parts \cite{Yadav1}. Texts spread across multiple lines are also removed. For texts that occur more than once, only a single copy is retained. Variations due to the archaeological context of the sites, stratigraphy, and type of object on which the texts are inscribed are, at present, not taken into account in the interests of retaining a reasonable sample size. 

The reasons for discarding texts which are spread over multiple lines are twofold. First, it is not clear if each line of a multi-line text is to be treated as a single text having a continuity of sequence, or if it is to be regarded as separate text. Second, assuming a continuity of sequence, the order in which the texts are to be read across lines remains ambiguous \cite{M77}. 

\begin{figure}
\includegraphics[width=8cm]{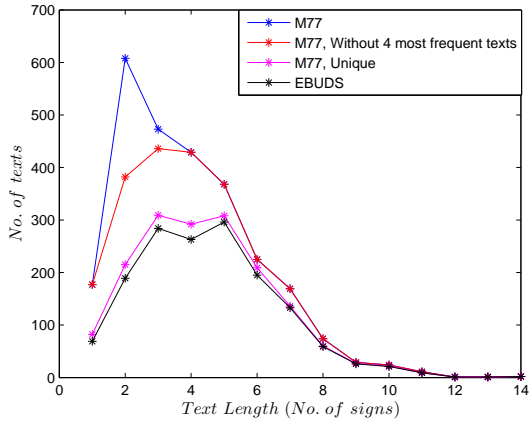}
\caption{\label{fig:TextLengthDistribution1}Text length distributions in the different corpora used in the analysis. The raw corpus (M77) contains four instances of outliers, texts of length $n=2$ and $n=3$ which occur in unusually large numbers. Keeping only single occurrences of these removes the sharp maximum around $n=2$ in the raw corpus. The corpus free of the outliers is then reduced again to keep only unique occurrences of the texts. This gives the M77-unique corpus. Finally, damaged, illegible and multi-line texts are removed to give the EBUDS corpus. Texts of length $n=3$ and $n=5$ are most frequent in this corpus.}
\end{figure}

The pruned EBUDS dataset contains $1548$ texts. In EBUDS, $40$ signs out of $417$ present in the sign list of M77 do not make their appearance. Out of these removed $40$ signs, one sign ($374$) appears $9$ times, one sign ($237$) appears $8$ times, two signs ($282$, $390$) appear $3$ times, three signs ($324$, $376$, $378$) appear twice and thirty-three signs appear only once in M77. Fig.~\ref{fig:TextLengthDistribution1} compares the distribution of texts lengths in various datasets and shows the effect of the removal of texts on the final dataset (EBUDS). We have already shown in our earlier work \cite{Yadav1}, that the relative frequency distribution of signs in EBUDS is comparable to M77 and hence EBUDS is a true representation of M77, with a reduction in total sign occurrences, but not in the percentage of total sign occurrences.

\begin{figure}
\center
\includegraphics[width=8cm]{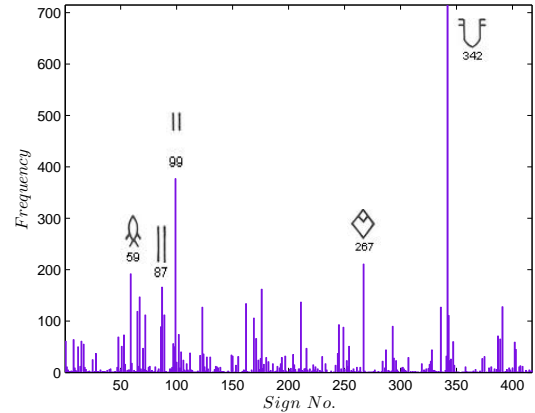}
\caption{\label{fig:SignFrequencyDistribution1}The frequency distribution of individual signs in the EBUDS corpus. The five most common signs are shown alongside the frequency bars. The relative frequency distribution does not change significantly between EBUDS and M77 corpora.}
\end{figure}

We first present the results of an empirical statistical analysis of the EBUDS corpus.  Fig.~\ref{fig:SignFrequencyDistribution1} shows the frequency distribution of signs in EBUDS. The sign corresponding to $342$ in M77 (\includegraphics[scale=.5]{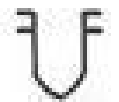}), is the most frequent sign, followed by signs $99$ (\includegraphics[scale=.5]{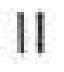}), $267$(\includegraphics[scale=.5]{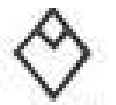}), and $59$ (\includegraphics[scale=.5]{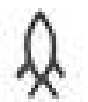}). The relative frequencies have no significant change in the M77 and EBUDS corpora.  

\subsection{Zipf-Mandelbrot law}

\begin{figure}
\center
\includegraphics[width=8cm]{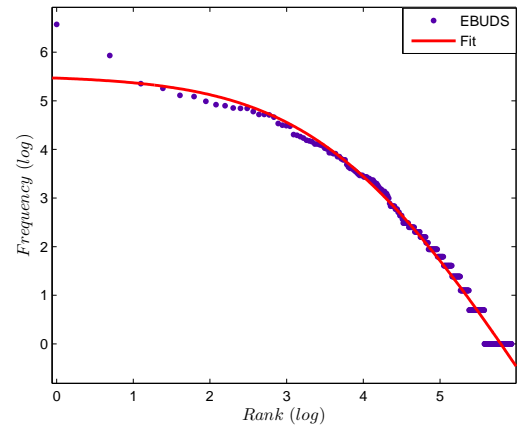}
\caption{\label{fig:EBUDSZipfsPlot1}The rank-ordered frequency distribution of signs $f_r$ plotted against the rank $r$ for the EBUDS corpus. The data is fitted well by the Zipf-Mandelbrot law, $\log f _r= a - b\log(r + c)$. For $c=0$ and $b=1$, this reduces to Zipf's Law, $f_r = a/r$. Both these laws are used to fit the frequency distribution of words in linguistic corpora.
Our fitted values are $a=15.39$, $b = 2.59$ and $c=44.47$. For English (the Brown Corpus), $a=12.43$, $b = 1.15$ and $c=100$ \cite{Schutze}.}
\end{figure}

The same data can be plotted in a different way, as a rank-ordered frequency distribution. The most frequent sign is given rank $r=1$ and its frequency is denoted by $f_1$, the next most frequent sign is given rank $r=2$ and its frequency is denoted as $f_2$ and so on, till all signs are exhausted. The rank-ordered frequency $f_r$ is then plotted against the rank $r$, on double logarithmic axes, as shown in  Fig.~\ref{fig:EBUDSZipfsPlot1}.  Remarkably, the data can be fit very well to the Zipf-Mandelbrot law, $\log f _r= a - b\log(r + c)$ \cite{Schutze}. The Zipf-Mandelbrot law is commonly used to fit the rank-ordered frequency distribution of {\it words} in corpora of {\it natural} languages. Qualitatively, a distribution which follows the Zipf-Mandelbrot law has a small number of tokens (words, signs) which contribute to the bulk of the distribution, but also a large number of rare signs which contribute to a long tail. To emphasise this point, it is estimated that English has approximately a million words, though a college graduate might know only between $60,000$ to $75,000$ of these, and yet be a completely competent user of the language \cite{Crystal}. The Indus script seems to follow the same pattern, with a small number of signs accounting for the majority of use, but with a large number of signs which are used infrequently. 

\subsection{Cumulative distribution}
\begin{figure}
\center
	\includegraphics[width=8cm]{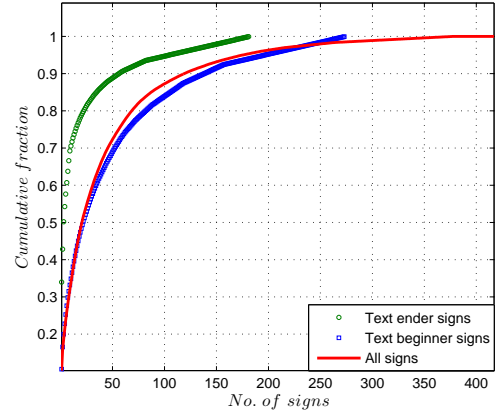}
	\caption{\label{fig:EBUDSCumGraph1}The cumulative frequency distribution of all signs, only text beginners, and only text enders in the EBUDS corpus. Approximately $69$ signs account for $80\%$ of the corpus. The script thus has a large number of signs which are used infrequently. The cumulative distributions for text beginners and text enders show an asymmetry, with only $23$ signs accounting for $80\%$ of all text enders, while one needs to take $82$ signs to account for $80\%$ of text beginners. This is clear evidence of the directionality in the sign usage.}
\end{figure}

To further follow up this point, we plot the cumulative frequency distribution of the signs in EBUDS in Fig.~\ref{fig:EBUDSCumGraph1}. As can be seen from the graph about $69$ signs account for $80 \%$ of the total data and the most frequent sign ($342$) alone accounts for $10\%$ of the complete EBUDS. This observation is consistent with previous analysis by Mahadevan for M77 corpus \cite{M77}. In the same graph, we plot the cumulative distribution of text beginners, that is the sign beginning the text, and text enders, that is the sign ending the text. Here, an interesting asymmetry is evident. Approximately $82$ text beginners account for $80\%$ of the text beginner usage, but only $23$ text enders are needed to account for the same percentage of text ender usage. Since the possible number of text beginners and text enders can be any of the $417$ signs, the numbers above indicate that both text beginners and text enders are well-defined, with text enders being more strictly defined than text beginners. This fact alone indicates that there is directionality in the use of the signs in Indus texts. Note that this conclusion of directionality is independent of any assumed direction of reading the sign sequence. Assuming the texts are read left to right would, instead, imply that text {\it beginners} are more strictly defined than text enders, again implying directionality. 

The analysis above has only been concerned with frequency distributions of single signs. We may extend the analysis to pairs of signs, triplets of signs and so on, as done by some of us earlier \cite{Yadav1, Yadav2}. This allows us to explore the very important feature of order and correlations between the signs, which are the outward manifestations of syntax. Syntax can be recognised even when the semantics is doubtful. For example, Chomsky's famous sentence ``Colourless green ideas sleep furiously'' becomes syntactically incorrect when the word order is changed to give, say, ``Green furiously ideas colourless sleep''. We recognise the first sentence as syntactically correct English, but of doubtful semantic value, but the second sentence is immediately recognised as syntactically invalid, even before we pause to think of its semantic content. It is this feature of the sign sequence of the Indus script, word order and correlation as an indicator of syntax, that we want to explore in detail, and which forms the main body of the present work. Such features are best explored in the $n$-gram model to be presented in the next section. 

\subsection{Probability of sequences}

Before exploring $n$-gram models, we briefly mention what sort of frequency distribution we should expect in the {\it absence} of correlations and also establish the notation we will be using for the remaining part of the paper. We denote a sequence of $N$ signs by ${\mathcal S}_N = s_1s_2\ldots s_N$, where each $s_i$ is one of the $417$ possible Indus signs. Each of these ${\mathcal S}_N$ is what we have been referring to as a text. The EBUDS corpus contains texts of maximum length $N=14$. By our analysis above, we have obtained frequency distributions for the signs $s_i$, by counting the number of times $c(s_i)$ that sign $s_i$ occurs in the corpus, and then dividing it by the total size of the corpus. This is identified with the {\it probability}, in the sense of maximum likelihood, of seeing the sign $s_i$ in a text,
\begin{equation}
P(s_i) = {c(s_i)\over \sum_i c(s_i)}
\end{equation}
In the absence of correlations, the joint probability that we see sign $s_2$ after sign $s_1$ is independent of $s_1$, and is just the product of their individual probabilities
\begin{equation}
P(s_1s_2) = P(s_1)P(s_2)
\end{equation}
Generalising, the probability of the string ${\mathcal S}_N = s_1s_2\ldots s_N$ is simply a product of the individual probabilities
\begin{equation}
P({\mathcal S}_N) = P(s_1s_2\ldots s_N) = P(s_1)P(s_2)\ldots P(s_N)
\end{equation}
In the absence of correlations, then, we have a generalisation of die throwing, where instead of $6$ possible outcomes, we have $417$ possible outcomes in each throw, and the $i^{th}$ outcome has a probability $P(s_i)$. Each throw outputs a single sign, and the outcome of a throw is independent of all previous throws. In $N$ throws, we generate a line of text ${\mathcal S}_N$. In probability theory, this process is called a Bernoulli scheme. In the next section, we will show that this is not adequate to model the sequence of Indus signs, indicating that there are significant correlations. We can then expect that an arbitrary permutation of  {\it signs} in an Indus text will produce a syntactically invalid sequence.

\begin{figure}[t]
\center
	\includegraphics[width=8cm]{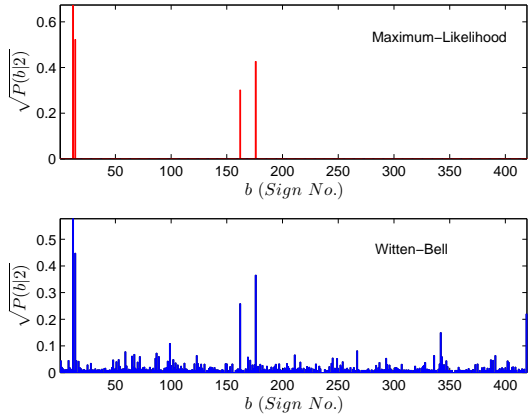}
		\caption{\label{fig:probBGiven2}The conditional probability $P(b|a=2)$ from the maximum likelihood estimate (above) and from Witten-Bell smoothing (below). The maximum likelihood estimate assigns zero probabilities to unseen sign pairs and results in a non-ergodic Markov chain. The Witten-Bell smoothing algorithm reduces the probabilities of the seen sign pairs and distributes the reduction over unseen sign pairs. This gives an ergodic Markov chain. The square root of conditional probabilities are plotted in each case to highlight the probabilities of unseen sign pairs.}
\end{figure}

\section{$n$-gram model for the Indus script} 
\label{sec:ngrams}

The main purpose of an $n$-gram model is to capture the correlations that exist between tokens $s_1, \ldots ,s_N$ in a sequence ${\mathcal S}_N$. Thus, conditional probabilities form the core of an $n$-gram model. Specifically, for a string ${\mathcal S}_N = s_1s_2\ldots s_N$ the $n$-gram model is a specification of conditional probabilities of the form $P(s_N|s_1s_2\ldots s_{N-1})$, quantifying the probability that the previous $N-1$ signs of the sub string ${\mathcal S}_{N-1}=s_1s_2\ldots s_{N-1}$ is followed by the sign $s_N$. Given the $n$-gram conditional probability, and the relation between joint and conditional probabilities $P(ab) = P(b|a)P(a)$,  the probability of the string ${\mathcal S}_N$ can be written as,
\begin{eqnarray}
P({\mathcal S}_N) &=& P(s_N|s_1\ldots s_{N-1})P(s_1\ldots s_{N-1})\\\nonumber
&=&P(s_N|{\mathcal S}_{N-1})P({\mathcal S}_{N-1})
\end{eqnarray}
Recursively applying $P(ab) = P(b|a)P(a)$ to the leftmost terms, we obtain the probability as a product over conditional probabilities 
\begin{equation}
P({\mathcal S}_N) = \prod_{i=i}^N P(s_i|{\mathcal S}_{i-1})
\end{equation}
In the above, it is understood that $\mathcal{S}_0$ is a special token indicating the start of the string. Note that the above expression is an identity that follows from the basic rules of probability and contains no approximations. As an example, the probability of a string of length three, $s_1s_2s_3$, is given as a product of trigram, bigram and unigram probabilities
\begin{equation}
P(s_1s_2s_3) = P(s_3|s_1s_2)P(s_2|s_1)P(s_1)
\end{equation}
Clearly, for an $n$-gram model to be tractable, only a finite number of such probabilities can be retained. In the simplest bigram model, all correlations beyond the preceding sign are discarded, so that
\begin{equation}
P(s_N|s_1\ldots s_{N-1}) = P(s_N|s_{N-1})
\end{equation}
In a trigram model, all correlations beyond two preceding signs are discarded, so that
\begin{equation}
P(s_N|s_1\ldots s_{N-1}) = P(s_N|s_{N-1}s_{N-2})
\end{equation}
In a general $n$-gram model, all correlations beyond the $(n-1)$ preceding signs are discarded. An $n$-gram model can then be thought of as an $(n-1)^{th}$ order Markov chain in a state space consisting of the signs $s_i$. The introduction of the Markov assumption is the main approximation in $n$-gram models. $N$-grams were first used by Markov to analyse the probability of a consonant following a vowel in Pushkin's {\it Eugene Onegin}. Shannon applied word $n$-grams to model sentences in English text and with $n=6$ obtained sentences which have a remarkable syntactic similarity to English. Since then, $n$-gram models have found wide use in many fields where sequences are to be analysed, including bioinformatics, speech processing and music. 

The theory and applications of $n$-grams are described in several excellent textbooks \cite{Schutze,Martin} and need not be repeated here. We summarise below our method of obtaining probabilities from counts, the use of smoothing and backoff to account for unseen $n$-grams, the measures we use for evaluating the $n$-gram model, and finally the tests we use to assign a statistical significance to the correlations. 

$N$-gram probabilities are obtained from counts. For single signs, the counts are unambiguous. However, for sign pairs, it is possible that a rare combination may not be present in a small sized corpus. The count, and the resulting probability estimate for that sign pair, then is zero. However, in the absence of reasons to the contrary, common sense dictates that no sign pair should have a strictly zero probability. This intuition is quantified by various rules which remove probability weight from seen sign pairs and distributes it to sign pairs which have never been seen, but are nonetheless not impossible. Common amongst such ``smoothing'' rules are Laplace's add-one rule,  a method developed by Turing and Good in breaking the Enigma code, and a more recent algorithm due to Witten and Bell \cite{Chen}. Here, we use the Witten-Bell algorithm to smooth our $n$-gram models. In Fig.~\ref{fig:probBGiven2} we show the estimate of the probability of a sign being followed by sign $2$, $P(b|2)$ before smoothing and after Witten-Bell smoothing. In above panel, the only non-zero probabilities are those corresponding to signs $12, 14, 162,$ and $176$. These probabilities sum to one, indicating that other sign pairs are impossible. The Witten-Bell smoothing algorithm restores a finite, but small probability to the unseen sign pairs, ensuring again that all probabilities sum to one. Apart from being a more reasonable way of estimating probabilities from counts, it also ensures that the resulting Markov chain is ergodic\addtocounter{footnote}{0}\footnote{A Markov chain is ergodic if it is possible to reach every state from every state (not necessarily in one move).}. An ergodic Markov chain is essential in such applications, since otherwise, probabilities of all strings containing  unseen $n$-grams vanish.

Smoothing is not the only way of correcting for unseen $n$-grams. Another method, called backoff uses probabilities of $(n-1)$-grams to estimate $n$-gram probabilities. Thus, the probability of unseen trigrams can be estimated from that of seen bigrams and unigrams. Here, we used the Katz backoff algorithm to estimate bigram, trigram and quadrigram probabilities when appropriate. 

The estimation of $n$-gram probabilities from $n$-gram counts is called learning. A learned $n$-gram model can then be tested to see if it produces $n$-grams in the corpus. To avoid circularity, the corpus is usually divided into a training set, from which the probabilities are estimated, and a test set, on which the model is evaluated. There are several standard measures for evaluating the goodness of an $n$-gram model. Here, we use a standard measure, the  perplexity, which is related to the information theoretic cross-entropy. We also do a cross-validation test using standard procedures.

Finally, we need tests of association to ascertain the significance of sign pairs which appear more or less frequently than what would be predicted by the Bernoulli scheme model. For this, we use a log-likelihood test, testing the null hypothesis that there is no association between sign pairs. 

In the next section, we provide results obtained from an $n$-gram model of the Indus script. The analysis was done using the SRILM \cite{Stolcke} and NSP \cite{Ted} toolkits.

\section{$n$-gram analysis of the corpus}
\label{sec:bigram}

In this section we present the $n$-gram probablities obtained from the EBUDS corpus and draw inferences on the structure of the script from an analysis of the $n$-gram probabilities. 

We introduce, following usual practice, two additional tokens \texttt{<s>} and \texttt{</s>} which indicate the beginning and end of a text respectively. By convention, the unigram probability for the start token is unity, $P(\texttt{<s>})=1$, since every text must begin with \texttt{<s>}. The probability of  sign $a$ being a beginner is then $P( \texttt{<s>}a) = (P(a|\texttt{<s>})$, since $P(\texttt{<s>})=1$. The probability of  sign $a$ being an ender is $P(a\texttt{</s>})$. It is also assumed, as usual,  that $n$-gram probabilities do not depend on the location of the signs in a text, or more formally, that the Markov chain is stationary. 

In any $n$-gram study, a maximum value of $n$ has to be chosen in the interest of tractability, beyond which correlations are discarded. This can be done in an empirical fashion, balancing the needs of accuracy and computational complexity, using measures from information theory which discriminate between $n$-grams models with increasing $n$ \cite{Schutze, Martin}, or by more sophisticated methods like the Akaike information criterion which directly provides an optimal value for $n$ \cite{Tong}. 

In previous work, some of us have shown that bigram and trigram frequencies in the EBUDS corpus differ significantly from frequencies expected from a Bernoulli scheme \cite{Yadav1}. The small size of the corpus limits the ability to assess significance of quadrigrams and beyond, when using the method in \cite{Yadav1}. In our subsequent work \cite{Yadav2} it has been shown that $88\%$ of the texts of length 5 and above can be segmented using frequent unigrams, bigrams, trigrams and quadrigrams and complete texts of length 2, 3 and 4.  Moreover, frequent bigrams or texts of length 2 alone account for $52\%$ of the segmented corpus. Thus the bulk of the corpus can be segmented with $n$-grams with $n$ not exceeding $4$, and almost half the corpus can be segmented into bigrams alone. This indicates that correlations beyond quadrigrams have relatively less significance. 

Here, we use an information theoretic measure, the perplexity, which we explain in detail below, to estimate an optimal value for $n$. Summarising the results of this analysis, which is presented in Table.~\ref{tab:Perplexity}, we find that the perplexity monotonically decreases as $n$ ranges from $1$ to $3$ (corresponding to unigram, bigram and trigram correlations), but then saturates beyond $n=4$ (corresponding to quadrigram and higher correlations). This confirms the observation made in \cite{Yadav2} that pentagram and higher correlations are of relatively less significance in the EBUDS corpus. From the differential reduction in perplexity with increase in model order, it is clear that the most significant correlations are due to bigrams, with somewhat modest trigram correlations, and almost negligible quadrigram correlations. Our subsequent analysis is, therefore, wholly based on bigrams. The main conclusions that we draw in this paper on the structure of the script are expected to remain broadly unaltered with the inclusion of trigram and quadrigram correlations. We expect a modest improvement in the predictive power of the $n$-gram model when trigram and quadrigram correlations are included. The role of higher-order correlations will be more fully explored in forthcoming work.

\begin{table}[t]
\caption{\label{tab:Significance}Significant sign pairs from the log-likelihood (LL) measure of association for bigrams. The $20$ most frequent sign pairs (first column) are compared with the $20$ most significant sign pairs (third column). The most frequent sign pairs are not necessarily the most significant sign pairs, as measured by the log-likelihood measure of association.}
\centering
\begin{tabular}{lcc|lcc}
\hline
Sign Pair&Rank& Frequency & Significant& Rank & LL Value\\
&(Naive)&(EBUDS)& Sign Pair&(LL)&\\
\hline
$267, 99$&$1$&$168$&$267, 99$&$1$&$792.40$\\
$336, 89$&$2$&$75$&$336, 89$&$2$&$522.03$\\
$342, 176$&$3$&$59$&$342, 1$&$3$&$286.46$\\
$8, 342$&$4$&$58$&$51, 130$&$4$&$252.02$\\
$391, 99$&$5$&$56$&$342, 176$&$5$&$210.24$\\
$347, 342$&$6$&$56$&$347, 342$&$6$&$208.68$\\
$342, 1$&$7$&$48$&$8, 342$&$7$&$201.77$\\
$293, 123$&$8$&$40$&$293, 123$&$8$&$196.14$\\
$87, 59$&$9$&$39$&$245, 245$&$9$&$195.67$\\
$48, 342$&$10$&$38$&$130, 149$&$10$&$181.35$\\
$171, 59$&$11$&$36$&$171, 59$&$11$&$169.99$\\
$249, 162$&$12$&$34$&$249, 162$&$12$&$156.07$\\
$89, 211$&$13$&$34$&$391, 99$&$13$&$155.63$\\
$245, 245$&$14$&$33$&$222, 254$&$14$&$147.21$\\
$59, 211$&$15$&$31$&$182, 293$&$15$&$137.34$\\
$51, 130$&$16$&$27$&$150, 123$&$16$&$132.38$\\
$65, 67$&$17$&$27$&$89, 211$&$17$&$130.77$\\
$99, 67$&$18$&$26$&$216, 254$&$18$&$128.64$\\
$162, 342$&$19$&$25$&$171, 8$&$19$&$114.87$\\
$123, 343$&$20$&$25$&$87, 59$&$20$&$111.01$\\
\hline
\end{tabular}
\end{table}
\subsection{\label{sec:sub1}Inferences from bigram matrix}

We now present the results of a bigram based analysis of the sequence of signs in the EBUDS corpus. To remind ourselves, in the bigram model, it is assumed that the probability $P(s_N|s_1\ldots s_{N-1})$ depends only on the immediately preceding sign and is the same as $P(s_N|s_{N-1})$. The bigram model is fully specified by the unigram probabilities $P(s_i)$ and the bigram probabilities $P(s_i|s_{i-1})$. 

\begin{figure*}
\center
\includegraphics[width=15cm]{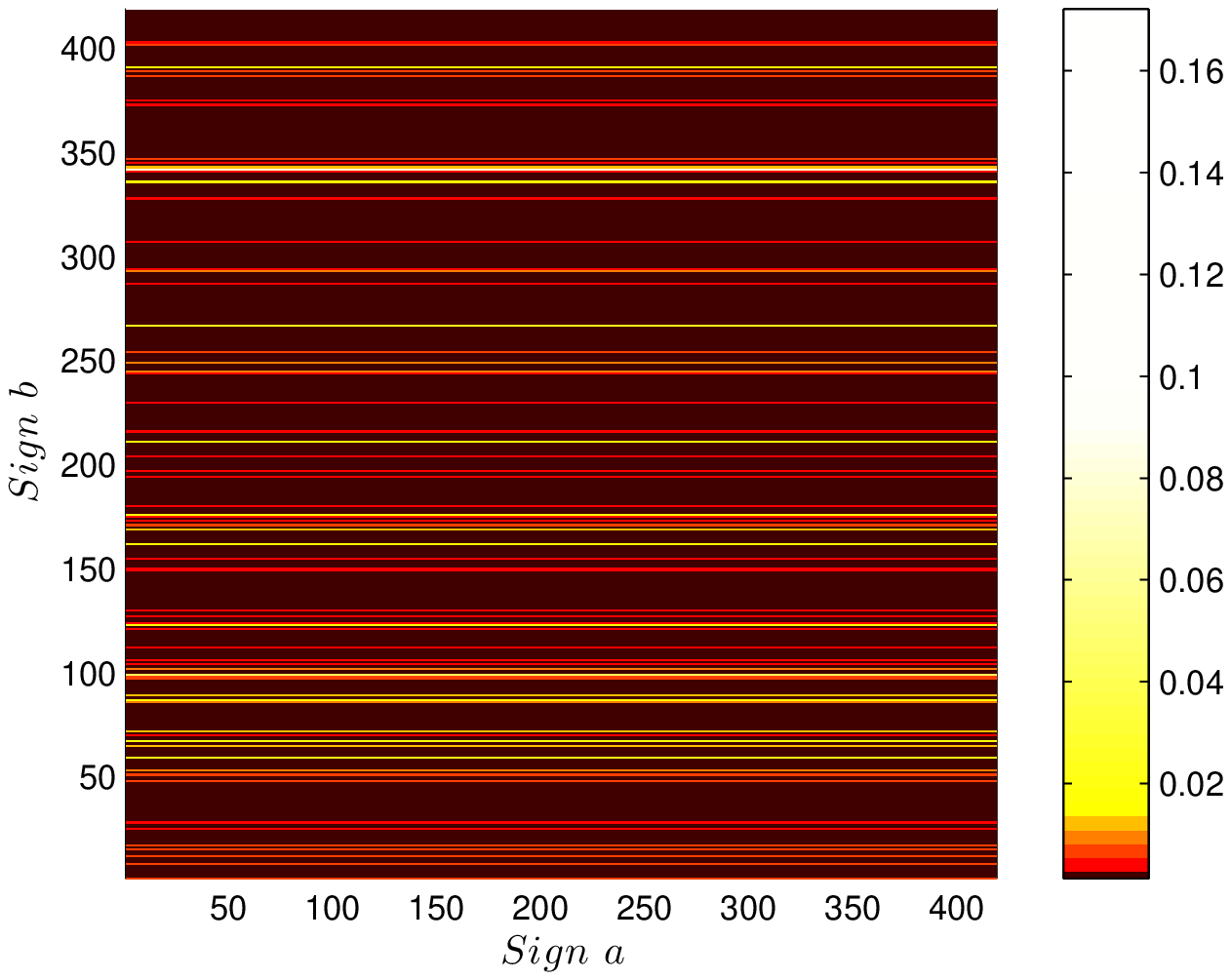}
\includegraphics[width=15cm]{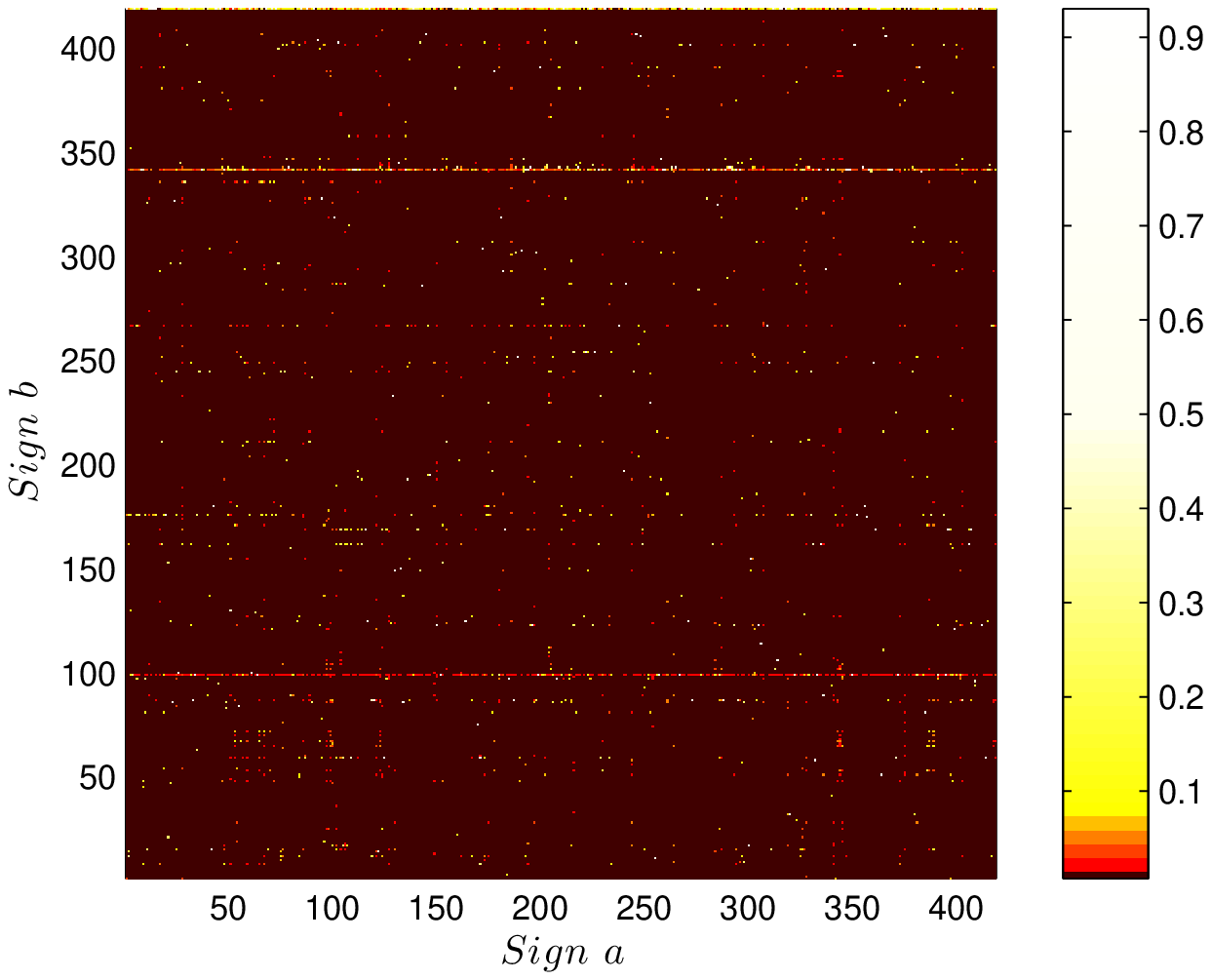}
\caption{\label{fig:CondProbMatrices}Bigram probability $P(b|a)$ for a random distribution with no correlations amongst the signs (above) and for the EBUDS corpus (below). Horizontal lines in the upper matrix imply that the conditional probability of a sign $b$ following a sign $a$ is equal to probability of sign $b$ itself. The bigram probability $P(b|a)$ after Witten-Bell smoothing is shown in the lower plot. The difference between the two matrices indicates the presence of correlations in the texts.}
\end{figure*}

We compare the bigram probabilities in the absence of correlations with the bigram probabilities for EBUDS corpus after Witten-Bell smoothing in the two plots of Fig.~\ref{fig:CondProbMatrices}. If there is no correlation between $b$ and $a$, we expect $P(b|a) = P(b)$, that is the conditional probability of $b$ is identical to the marginal probability. We show this marginal probability in the first plot of Fig.~\ref{fig:CondProbMatrices}. In the second plot of Fig.~\ref{fig:CondProbMatrices} we show the matrix of bigram probabilities $P(b|a)$ that sign $b$ follows sign $a$ in the corpus.  It is clear even by a casual observation, that there are significant differences between the two plots. This signals the presence of correlations in the script. 

\begin{figure*}[]
\center

   \includegraphics[width=8cm,bb=88 415 509 568,clip]{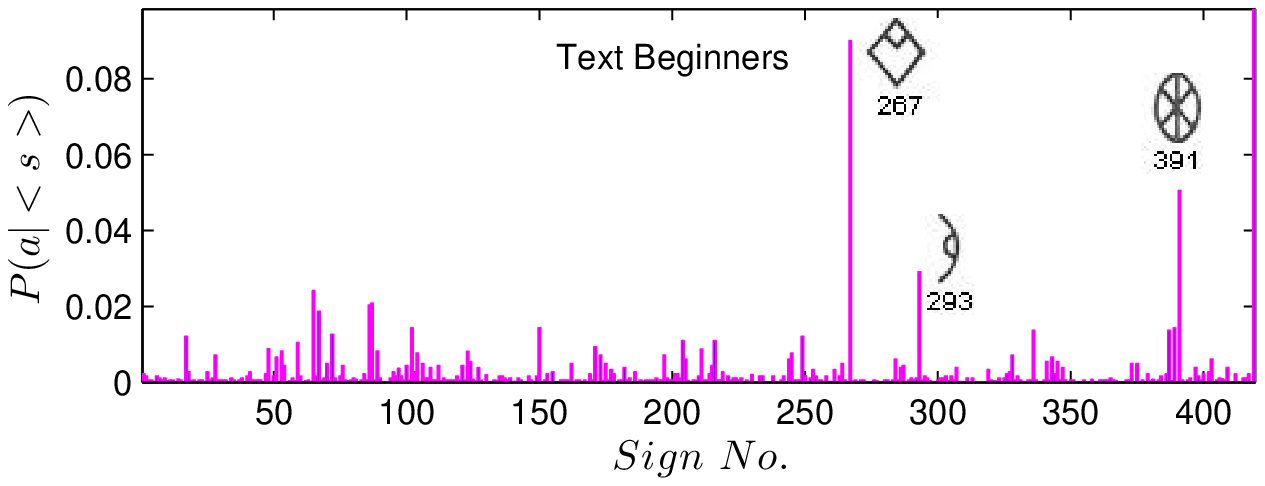}
  \includegraphics[width=8cm,bb=88 415 507 568,clip]{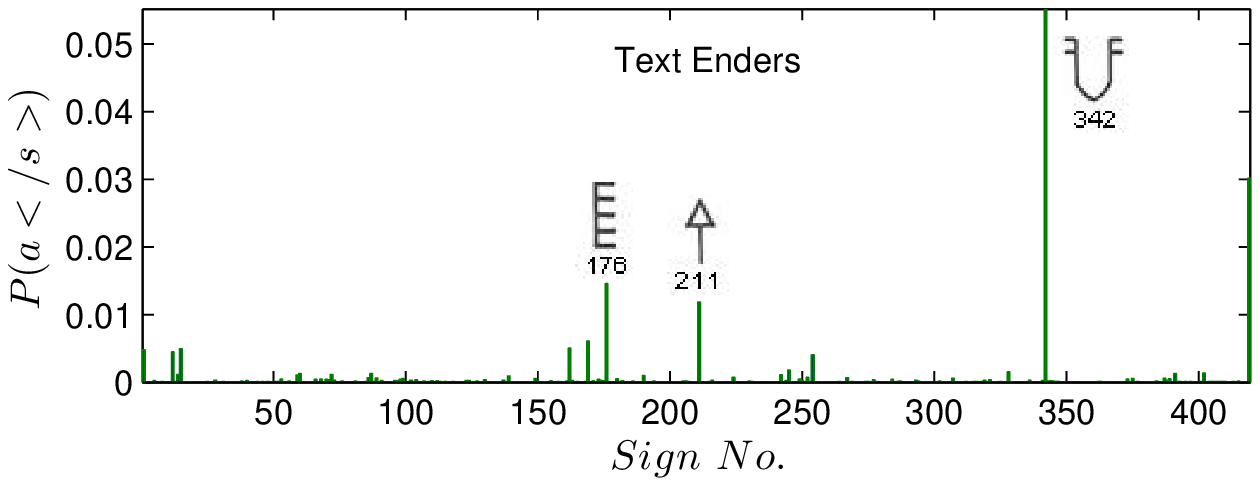} \caption{\label{fig:wbCondBegEnderPlots1WithSigns}Probability $P(a|\texttt{<s>})$ of a sign $a$ following the start token \texttt{<s>} (text beginners) and probability $P(a\texttt{</s>})$ of sign $a$ preceding the end token \texttt{</s>} (text enders) from bigram matrix $P(b|a)$ with Witten-Bell smoothing. Text beginners with a significant probability are more numerous than text enders at the same threshold of probability.}
  
	\includegraphics[width=8cm]{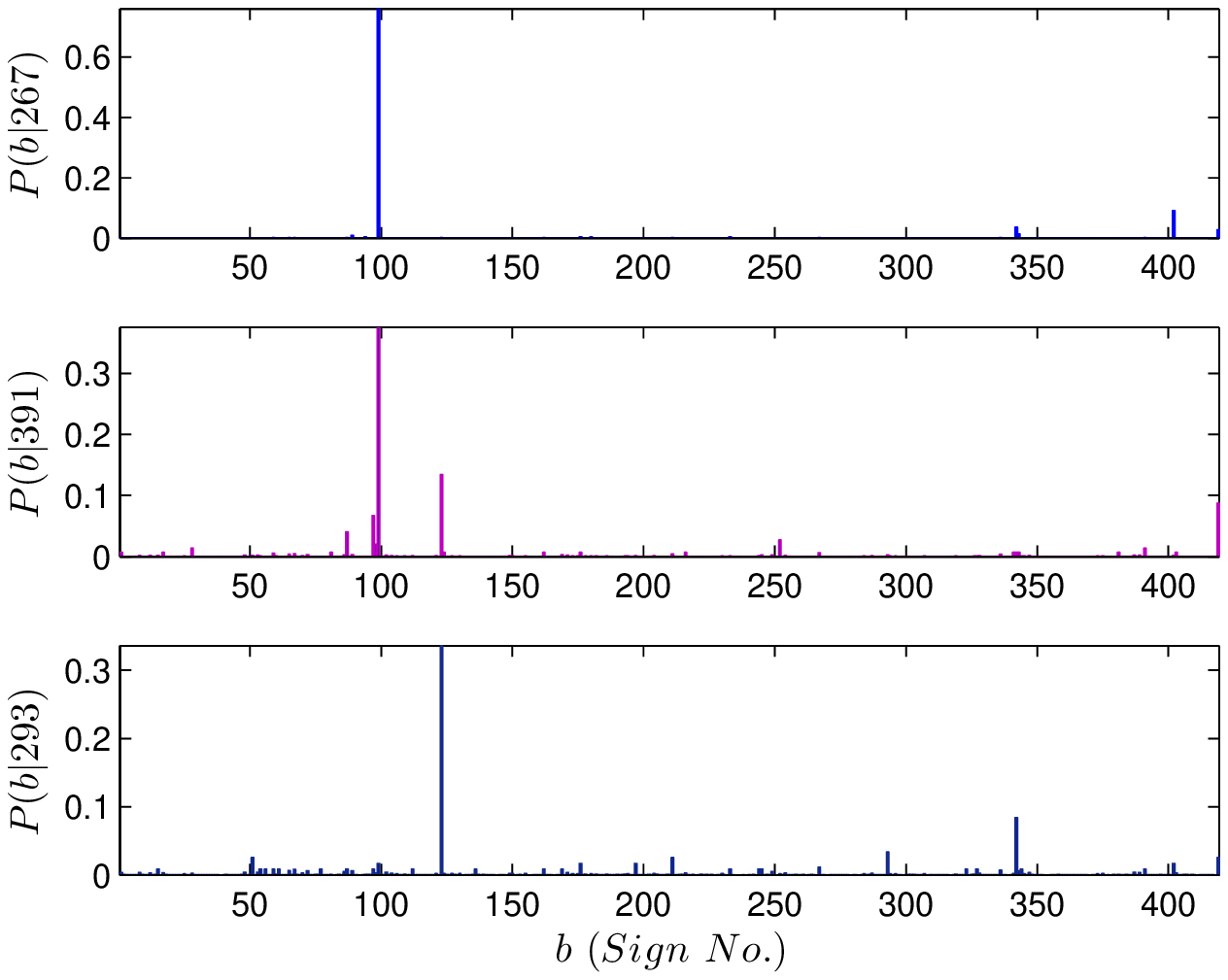}
	\includegraphics[width=8cm]{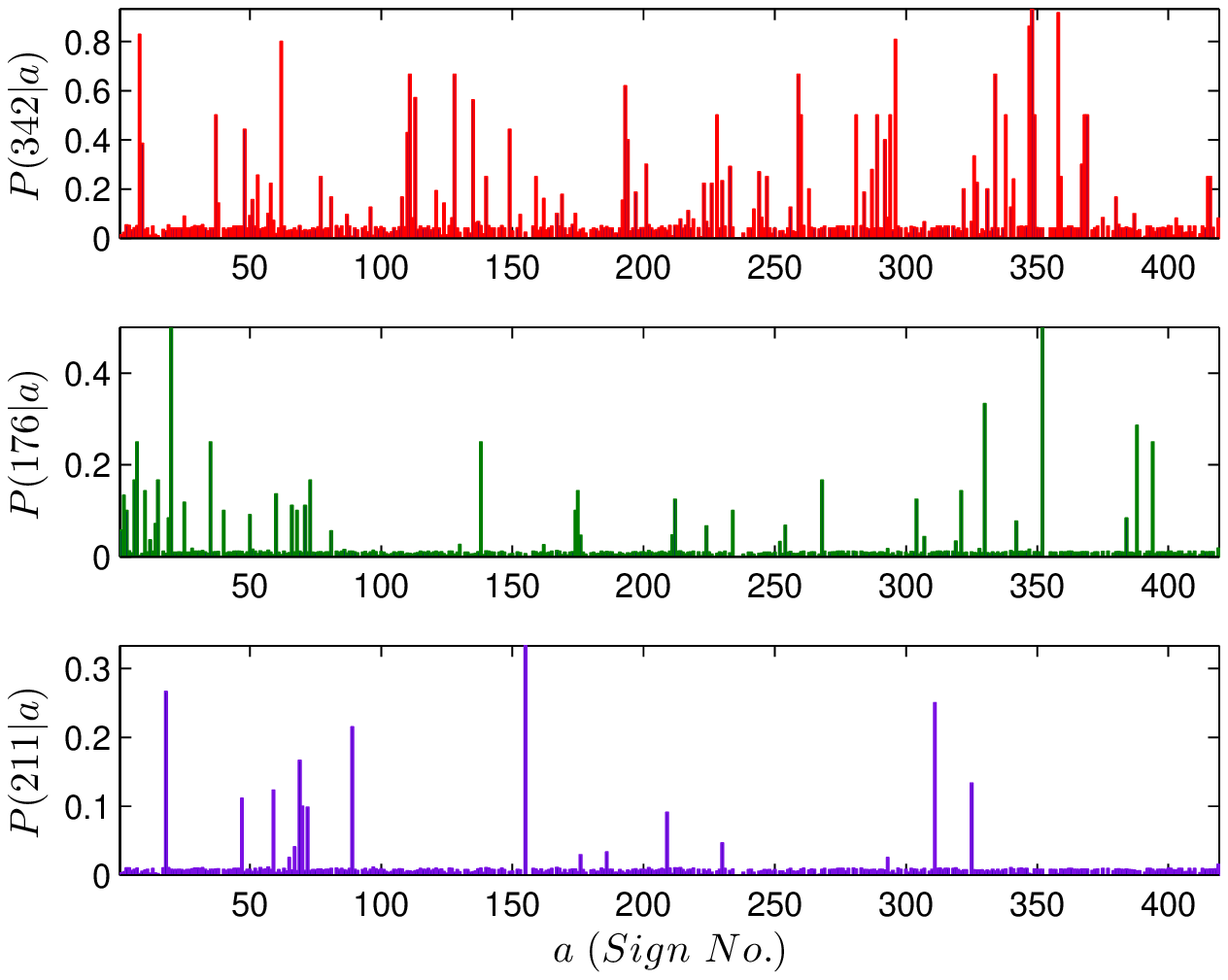}\caption{\label{fig:wbBegEnderAnalsisPlots12}Conditional probability plots for text beginners $a = 267$, $391, 293$ followed by sign $b$ and for texts enders $ b= 342$, $176, 211$ preceded by sign $a$  from bigram matrix $P(b|a)$ with Witten-Bell smoothing. Text beginners are more selective in terms of the number of signs which can follow them than text enders, which can have a large number of signs preceding them.}

\includegraphics[width=8cm,bb=88 415 509 568,clip]{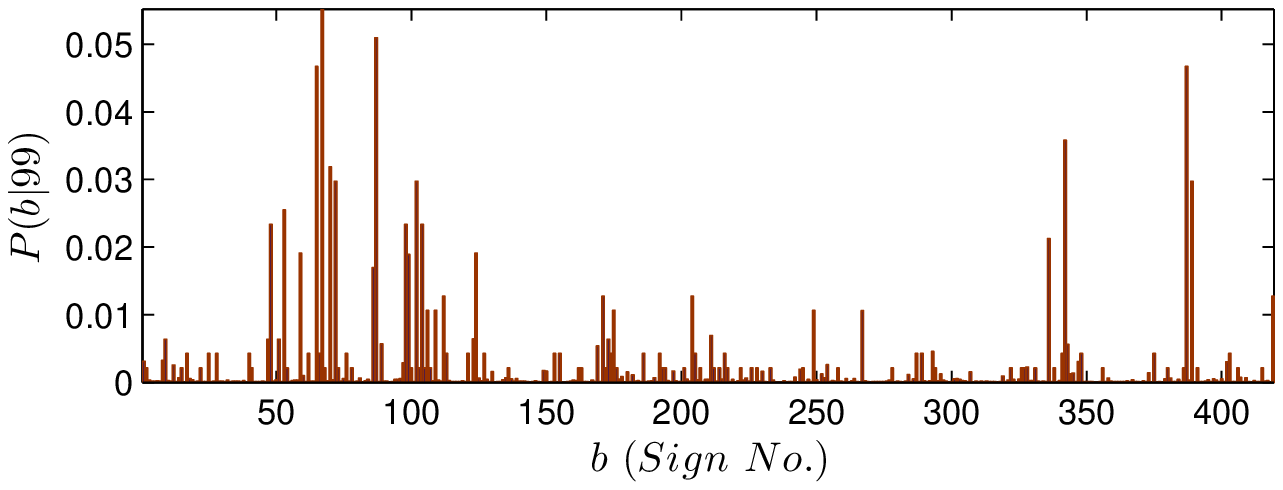}	
\includegraphics[width=8cm,bb=88 415 507 568,clip]{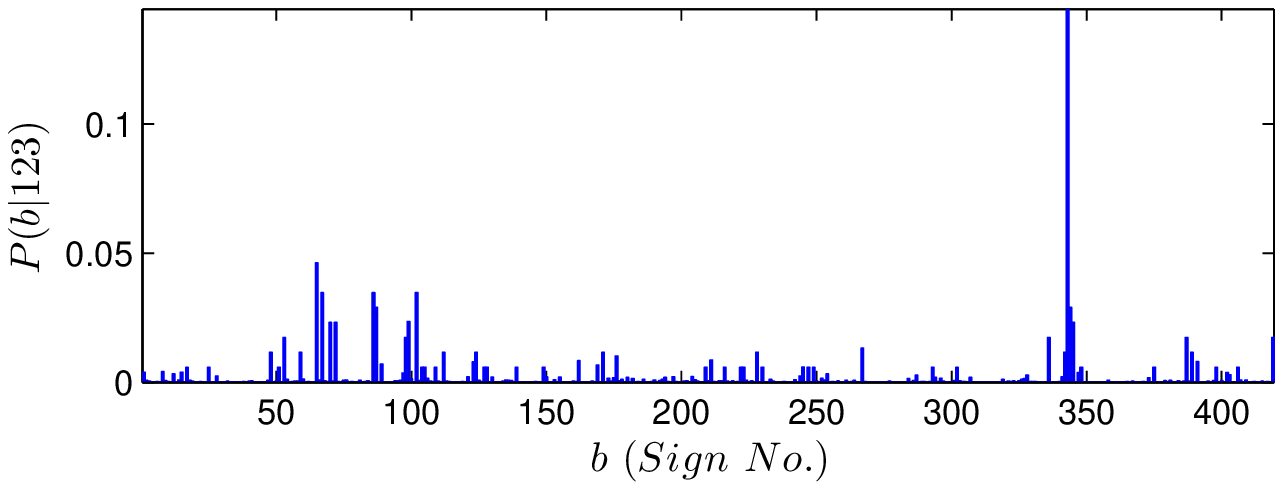}\caption{\label{fig:wbBegEnderAnalysisPlots3}Conditional probability plots for sign $b$ following text beginners $a = 99$ and $a = 123$. The number of signs following the signs $99$ and $123$ is greater than the  number of signs following text beginners $267$, $391$ and $293$ (Fig.~\ref{fig:wbBegEnderAnalsisPlots12}) .}
 	\end{figure*}
 	
In Fig.~\ref{fig:wbCondBegEnderPlots1WithSigns} we show the text beginner and text ender sign probability distributions. The more careful analysis using smoothed bigram probabilities confirms our earlier conclusion, based on a raw counts, that text enders are more strictly defined than text beginners. 

We can further analyse the nature of correlations of a sign with other signs preceding or following it using the results of bigram analysis. As an example, we explore the correlations of the three most frequent text beginners (sign numbers $267$, $391$ and $293$) and the three most frequent text enders (sign numbers $342$, $176$ and $211$) shown in Fig.~\ref{fig:wbCondBegEnderPlots1WithSigns} with other signs. It can be inferred from the plots  of conditional probabilities (Fig.~\ref{fig:wbBegEnderAnalsisPlots12}) i.e. $P(b|267)$, $P(b|391)$ and $P(b|293)$ for the text beginners and $P(342|a)$, $P(176|a)$ and $P(211|a)$ for the text enders, that the text beginners $267$, $391$ and $293$ are more selective in terms of the number of signs which can follow them in comparison to the text enders $342$, $176$ and $211$ which can be preceded by relatively larger number of signs. Thus, there is greater flexibility for signs preceding the text enders than the signs which tend to follow the text beginners. Moreover, if we further go into the details of the restricted number signs $99$, $123$ which follow the text beginners we find that the nature of the most frequent signs following the text  beginners is quite similar to the text enders (though in reverse direction), in terms of the number of signs which follow them as shown in the plots of $P(b|99)$ and $P(b|123)$ (Fig.~\ref{fig:wbBegEnderAnalysisPlots3}). In other words, the correlation between the text beginners and the signs following them is stronger than the correlation of these immediate neighbours of text beginners with the sign following them, a fact which helps us in finding the weaker and stronger junctions in the text \cite{ Yadav2}. The diagonal elements $P(b|b)$ of the matrix of bigram probabilities are the probabilities of sign pairs with same signs and the most frequent sign pairs with repeating signs in the corpus are ($153$, $153$) and ($245$, $245$).

\begin{table}[t]
\caption{\label{tab:Entropy}The entropy and mutual information of the EBUDS corpus. The entropy is smaller than a random equiprobable sequence of $417$ signs. The mutual information is non-zero, indicating the presence of correlations between consecutive signs.}
\centering
\begin{tabular}{ccc}
\hline
Measure& Random & EBUDS \\
\hline
\\
Entropy (H)&$8.7039$&$6.6811$ \\
\\
Mutual information (I) & $0$ &$2.236$\\
\\
\hline
\end{tabular}
\end{table}

 \begin{table}[]
\caption{\label{tab:Perplexity}Perplexity  and the $n$-gram cross entropy $H_n(Q,P)$ for the EBUDS corpus. The perplexity reduces dramatically when bigram correlations are included, has a small but significant reduction with trigram correlations, but then saturates beyond quadrigram correlations. This indicates that a bigram model is optimal for capturing the syntax in the EBUDS corpus.}
\centering
\begin{tabular}{c|ccccc}
\hline
$n$ & $1$ &  $2$ & $3$ & $4$ & $5$\\
\hline
\\
Perplexity ($\mathcal{P}$) & 68.82 & 26.69 & 26.09 & 25.26 & 25.26 \\
\\
$H_n(Q, P)$ & 6.10 & 4.74 & 4.71 & 4.66 & 4.66 \\
\\
\hline
\end{tabular}
\end{table}

\subsection{\label{sec:sub2}Test of significance}

To quantitatively assess the significance of the correlations, and to obtain the statistically significant sign pairs, we need to test the null hypothesis that signs $a$ and $b$ are independent. In Table~\ref{tab:Significance} we give the most frequent sign pairs as well as the ones which are most statistically significant. Here, we enumerate the $20$ most significant sign pairs on the basis of log-likelihood measure of association for bigrams. It is interesting to note the most significant sign pairs are not always the most frequent ones given in the first column of Table~\ref{tab:Significance} for comparison. 
\begin{table}[t]
\caption{\label{tab:GeneratedTexts}Examples of texts generated by the bigram model. The texts are to be read from the right to the left. Some of the texts generated by the model occur in the corpus.}       
        \centering
        \begin{tabular}{r}
        \hline\\
				\includegraphics[scale=.5]{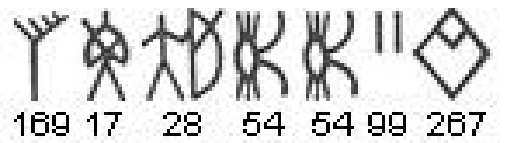}\\\\
        \newline 
        \includegraphics[scale=.5]{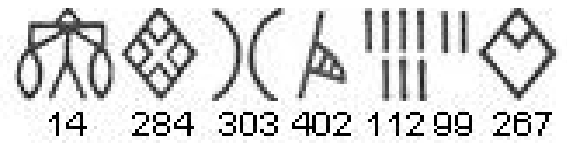}\\\\
        \newline
        \includegraphics[scale=.5]{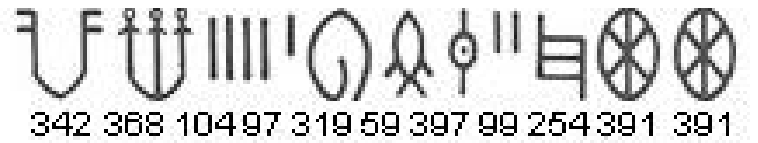}\\\\
        \newline
		    \includegraphics[scale=.5]{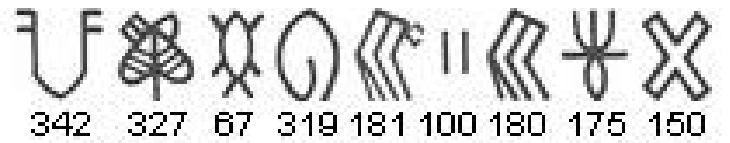}\\\\
        \newline
		    \includegraphics[scale=.5]{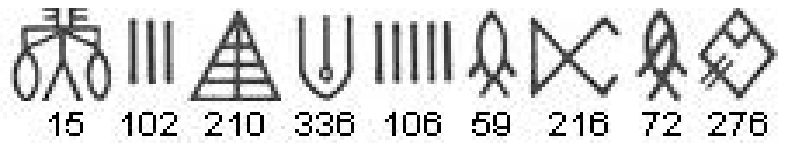}\\\\
		    \hline
        \end{tabular}
      
\end{table}

\subsection{\label{sec:sub3}Entropy, Mutual Information and Perplexity}

\begin{table*}[tp]
\caption{\label{tab:FillingTable}Suggested restoration of signs missing from texts. The last column lists the suggested restorations in decreasing order probability (Left to Right).}
\centering
\begin{tabular}{crrrc}
\hline\\
Text No. & Text & Incomplete Text & Most Probable Restoration & Probable Restored Sign\\\\
\hline\\
\newline
4312&\begin{minipage}{1.25in}\hspace{1.25cm}\includegraphics[scale=.5]{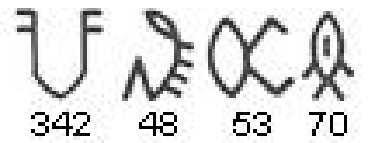}\end{minipage}&\begin{minipage}{1.25in}\hspace{1.4cm}\includegraphics[scale=.5]{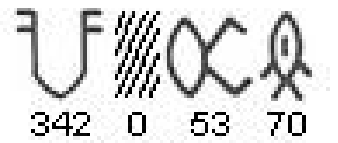}\end{minipage}&\begin{minipage}{1.25in}\hspace{1.25cm}\includegraphics[scale=.5]{4312}\end{minipage}&\begin{minipage}{1.2in}\hspace{6mm}\includegraphics[scale=.5]{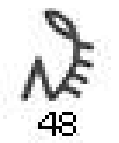}\includegraphics[scale=.5]{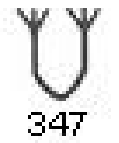}\includegraphics[scale=.5]{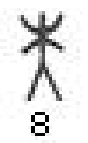}\end{minipage}\\\\

\newline
4016&\begin{minipage}{1.25in}\hspace{0.4cm}\includegraphics[scale=.5]{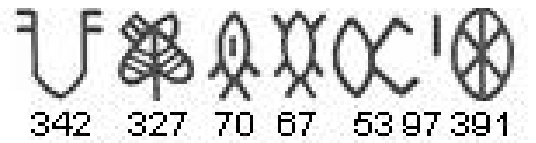}\end{minipage}&\begin{minipage}{1.25in}\hspace{0.4cm}\includegraphics[scale=.5]{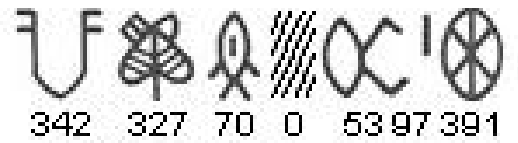}\end{minipage}&\begin{minipage}{1.25in}\hspace{0.4cm}\includegraphics[scale=.5]{4016}\end{minipage}&\begin{minipage}{1.25in}\hspace{8mm}\includegraphics[scale=.5]{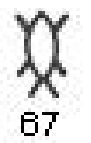}\includegraphics[scale=.5]{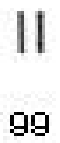}\includegraphics[scale=.5]{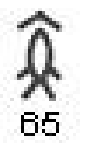}\end{minipage}\\\\

\newline
5237&\begin{minipage}{1.25in}\hspace{1cm}\includegraphics[scale=.5]{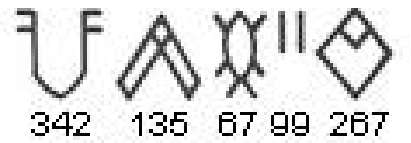}\end{minipage}&\begin{minipage}{1.25in}\hspace{1.10cm}\includegraphics[scale=.5]{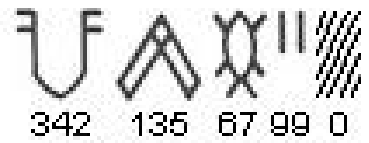}\end{minipage}&\begin{minipage}{1.25in}\hspace{1cm}\includegraphics[scale=.5]{5237}\end{minipage}&\begin{minipage}{1.25in}\hspace{8mm}\includegraphics[scale=.5]{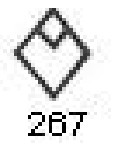}\includegraphics[scale=.5]{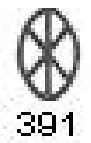}\end{minipage}\\\\

\newline
2653&\begin{minipage}{1.25in}\hspace{0.6cm}\includegraphics[scale=.5]{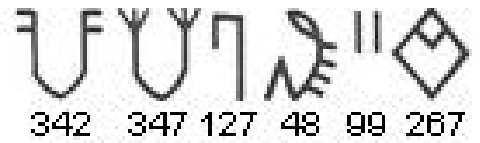}\end{minipage}&\begin{minipage}{1.25in}\hspace{0.8cm}\includegraphics[scale=.5]{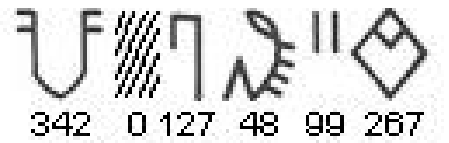}\end{minipage}&\begin{minipage}{1.25in}\hspace{0.6cm}\includegraphics[scale=.5]{2653}\end{minipage}&\begin{minipage}{1.25in}\hspace{8mm}\includegraphics[scale=.5]{347}\includegraphics[scale=.5]{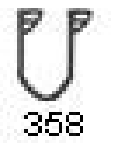}\includegraphics[scale=.5]{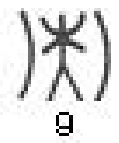}\end{minipage}\\\\

\newline
5073&\begin{minipage}{1.25in}\hspace{0.4cm}\includegraphics[scale=.5]{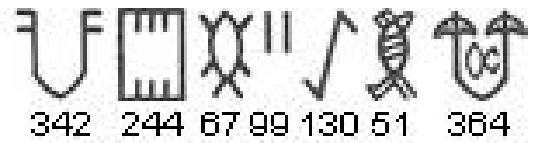}\end{minipage}&\begin{minipage}{1.25in}\hspace{0.5cm}\includegraphics[scale=.5]{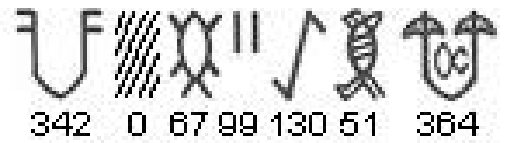}\end{minipage}&\begin{minipage}{1.25in}\hspace{0.4cm}\includegraphics[scale=.5]{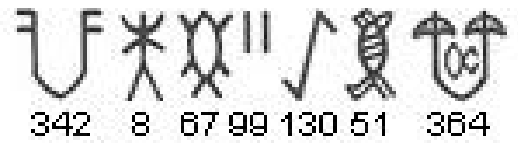}\end{minipage}&\begin{minipage}{1.25in}\hspace{8mm}\includegraphics[scale=.5]{8}\includegraphics[scale=.5]{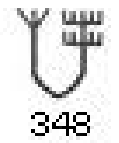}\includegraphics[scale=.5]{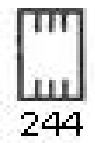}\end{minipage}\\\\

\newline
3360&\begin{minipage}{1.25in}\hspace{0.5cm}\includegraphics[scale=.5]{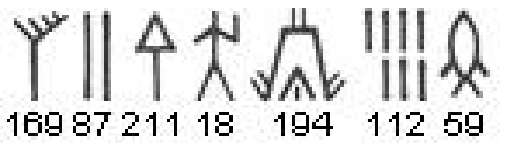}\end{minipage}&\begin{minipage}{1.25in}\hspace{0.6cm}\includegraphics[scale=.5]{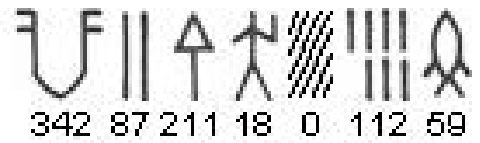}\end{minipage}&\begin{minipage}{1.25in}\hspace{0.5cm}\includegraphics[scale=.5]{3360}\end{minipage}&\begin{minipage}{1.25in}\hspace{8mm}\includegraphics[scale=.5]{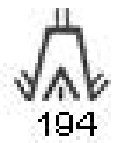}\end{minipage}\\\\

9071&\begin{minipage}{1.25in}\hspace{0.4cm}\includegraphics[scale=.5]{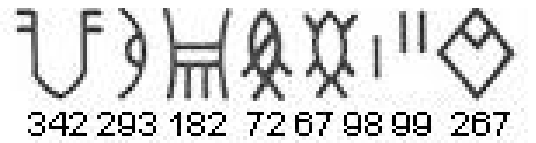}\end{minipage}&\begin{minipage}{1.25in}\hspace{0.3cm}\includegraphics[scale=.5]{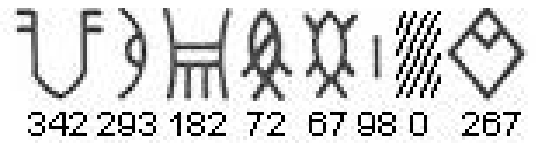}\end{minipage}&\begin{minipage}{1.25in}\hspace{0.4cm}\includegraphics[scale=.5]{9071}\end{minipage}&\begin{minipage}{1.25in}\hspace{8mm}\includegraphics[scale=.5]{99}\includegraphics[scale=.5]{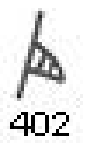}\includegraphics[scale=.5]{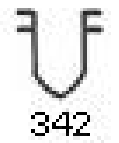}\end{minipage}\\\\

\newline
4081&\begin{minipage}{1.25in}\hspace{0.2cm}\includegraphics[scale=.5]{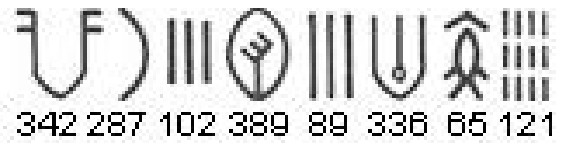}\end{minipage}&\begin{minipage}{1.25in}\hspace{0.20cm}\includegraphics[scale=.5]{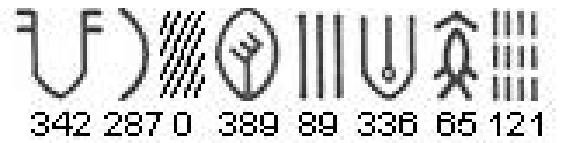}\end{minipage}&\begin{minipage}{1.25in}\hspace{0.20cm}\includegraphics[scale=.5]{4081}\end{minipage}&\begin{minipage}{0.25in}\hspace{2mm}\includegraphics[scale=.5]{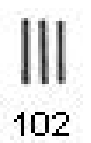}\end{minipage}\begin{minipage}{0.25in}\hspace{2mm}\texttt{</s>}\end{minipage}\begin{minipage}{0.25in}\hspace{2mm}\includegraphics[scale=.5]{65}\end{minipage}\\\\

\hline
\end{tabular}
\label{tab:gt}
\end{table*}

\begin{table*}[bp]
\caption{\label{tab:illegible}Suggested restoration of doubtfully read signs in the texts of M77 corpus. The last column lists the suggested restorations in decreasing order probability (Left to Right). The signs with asterisk sign at the top right are the doubtfully read signs which are being restored using the bi-gram model.}
\centering
\begin{tabular}{crrrc}
\hline\\
Text No. & Text & Incomplete Text & Most Probable Restoration & Probable Restored Sign\\\\
\hline\\
\newline
8302&\begin{minipage}{1.25in}\hspace{1cm}\includegraphics[scale=.5]{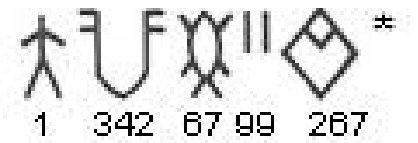}\end{minipage}&\begin{minipage}{1.25in}\hspace{1.30cm}\includegraphics[scale=.5]{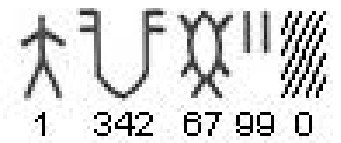}\end{minipage}&\begin{minipage}{1.25in}\hspace{1.2cm}\includegraphics[scale=.5]{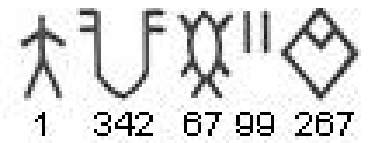}\end{minipage}&\begin{minipage}{1.25in}\hspace{7mm}\includegraphics[scale=.5]{267}\includegraphics[scale=.5]{391}\end{minipage}\\\\

\newline
5317&\begin{minipage}{1.25in}\hspace{1cm}\includegraphics[scale=.5]{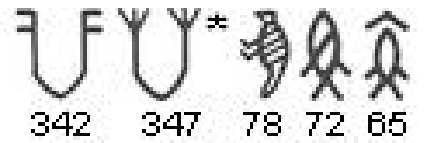}\end{minipage}&\begin{minipage}{1.25in}\hspace{1.2cm}\includegraphics[scale=.5]{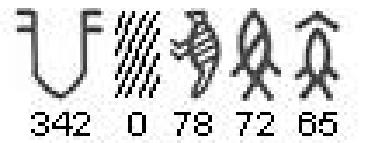}\end{minipage}&\begin{minipage}{1.25in}\hspace{1cm}\includegraphics[scale=.5]{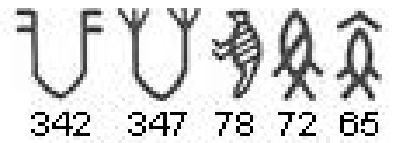}\end{minipage}&\begin{minipage}{1.25in}\hspace{7mm}\includegraphics[scale=.5]{347}\end{minipage}\\\\

\newline
1193&\begin{minipage}{1.25in}\hspace{2cm}\includegraphics[scale=.5]{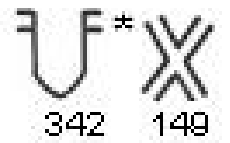}\end{minipage}&\begin{minipage}{1.25in}\hspace{2.3cm}\includegraphics[scale=.5]{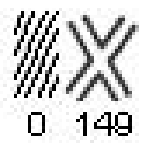}\end{minipage}&\begin{minipage}{1.25in}\hspace{2cm}\includegraphics[scale=.5]{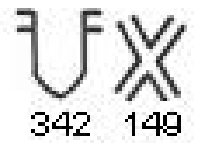}\end{minipage}&\begin{minipage}{1.25in}\hspace{7mm}\includegraphics[scale=.5]{342}\end{minipage}\\\\

\newline
1407&\begin{minipage}{1.25in}\hspace{1.1cm}\includegraphics[scale=.5]{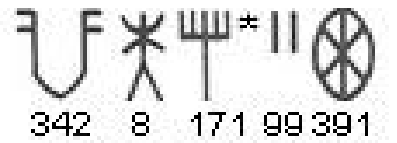}\end{minipage}&\begin{minipage}{1.25in}\hspace{1.3cm}\includegraphics[scale=.5]{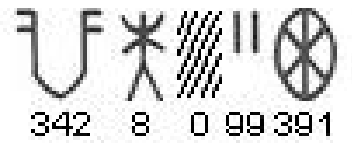}\end{minipage}&\begin{minipage}{1.25in}\hspace{1.2cm}\includegraphics[scale=.5]{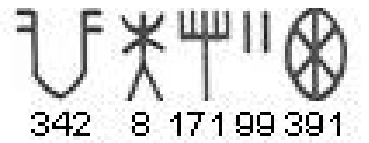}\end{minipage}&\begin{minipage}{1.25in}\hspace{7mm}\includegraphics[scale=.5]{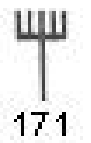}\includegraphics[scale=.5]{48}\includegraphics[scale=.5]{67}\end{minipage}\\\\

\newline
2179&\begin{minipage}{1.25in}\hspace{0.3cm}\includegraphics[scale=.5]{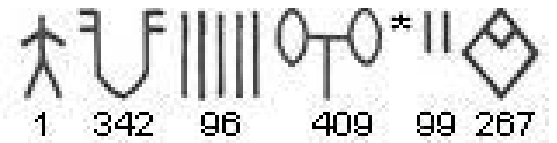}\end{minipage}&\begin{minipage}{1.25in}\hspace{0.8cm}\includegraphics[scale=.5]{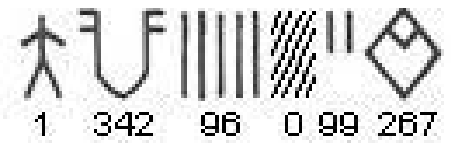}\end{minipage}&\begin{minipage}{1.25in}\hspace{0.4cm}\includegraphics[scale=.5]{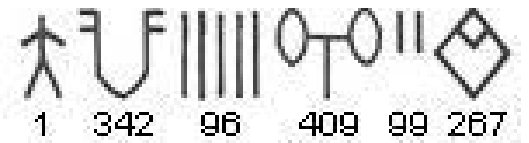}\end{minipage}&\begin{minipage}{1.25in}\hspace{7mm}\includegraphics[scale=.5]{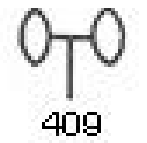}\end{minipage}\\\\

\newline
3396&\begin{minipage}{1.25in}\hspace{0.3cm}\includegraphics[scale=.5]{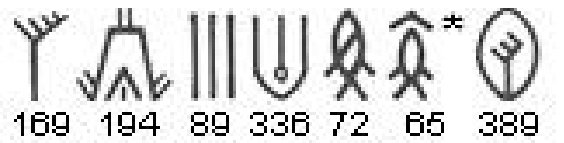}\end{minipage}&\begin{minipage}{1.25in}\hspace{0.5cm}\includegraphics[scale=.5]{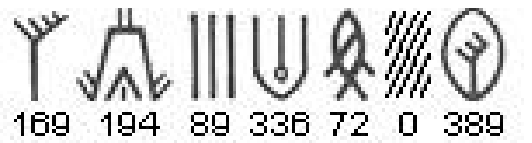}\end{minipage}&\begin{minipage}{1.25in}\hspace{0.4cm}\includegraphics[scale=.5]{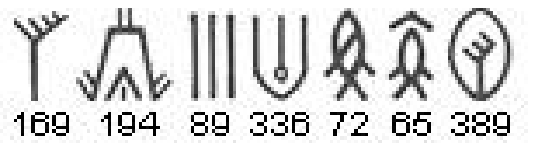}\end{minipage}&\begin{minipage}{1.25in}\hspace{7mm}\includegraphics[scale=.5]{65}\includegraphics[scale=.5]{67}\includegraphics[scale=.5]{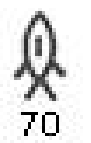}\end{minipage}\\\\

\newline
8101&\begin{minipage}{1.25in}\hspace{0.9cm}\includegraphics[scale=.5]{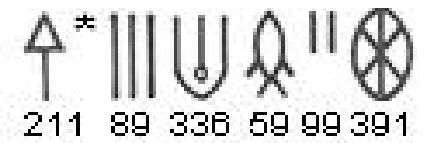}\end{minipage}&\begin{minipage}{1.25in}\hspace{1.1cm}\includegraphics[scale=.5]{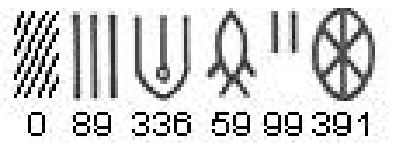}\end{minipage}&\begin{minipage}{1.25in}\hspace{1cm}\includegraphics[scale=.5]{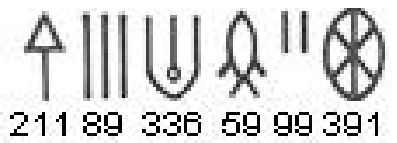}\end{minipage}&\begin{minipage}{1.25in}\hspace{7mm}\includegraphics[scale=.5]{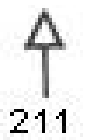}\end{minipage}\\\\

2802&\begin{minipage}{1.25in}\hspace{0.1cm}\includegraphics[scale=.5]{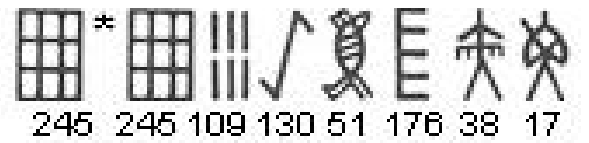}\end{minipage}&\begin{minipage}{1.25in}\hspace{0.4cm}\includegraphics[scale=.5]{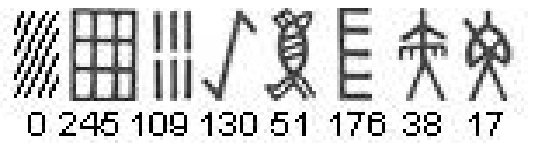}\end{minipage}&\begin{minipage}{1.25in}\hspace{0.1cm}\includegraphics[scale=.5]{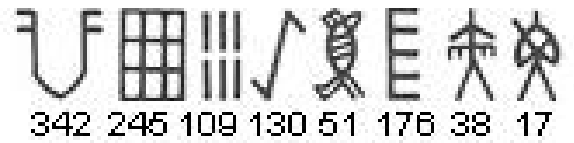}\end{minipage}&\begin{minipage}{0.25in}\includegraphics[scale=.5]{342}\includegraphics[scale=.5]{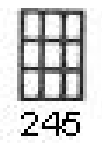}\end{minipage}\begin{minipage}{0.5in}\hspace{5mm}\texttt{</s>}\end{minipage}\\\\

\hline
\end{tabular}
\label{tab:gt}
\end{table*}

Information theoretic measures can also be used to make an estimate of the role of correlations. We use the informational theoretic entropy $H$ and the mutual information $I$, 
\begin{eqnarray}
H &=& -\sum_{a}P(a)\log_2 P(a),\\
I &=& \sum_{ab}P(ab)\log_2\left({P(ab)\over P(a)P(b)}\right),
\end{eqnarray}
to quantify the amount of correlation in the sequence.

In Table~\ref{tab:Entropy} we compare the entropy and mutual information of the corpus with that of a completely random sequence with no correlations. For this, the probability of signs is uniform $P(a) = 1/417$ and the joint probability is $P(ab) = P(a)P(b)$. This gives an entropy of $-\log(417)$ and a vanishing mutual information. In contrast, the entropy of the EBUDS corpus is $6.3$ and the mutual information is $2.3$. This indicates the presence of correlations, but the difference between the entropy and mutual information also indicates that there is flexibility in the sign usage, and the probability of a sign  is not completely determined by the preceding sign. 

A related information-theoretic measure, the cross-entropy, is useful in discriminating between $n$-gram models with different $n$. The main goal of $n$-gram analysis is to construct a good model for the probability of sequences in a corpus.  A metric which quantifies how close the estimated probability distribution is to the actual probability distribution is the cross-entropy. For a true distribution $Q(a)$ and its estimate $P(a)$, the cross entropy is defined as,
\begin{equation}
H(Q, P) = -\sum_a Q(a)\log_{2}P(a)
\end{equation}
The cross-entropy is minimum when the true and estimated probability distributions coincide,  $Q(a) = P(a)$. The perplexity $\mathcal{P}$, which is the measure commonly used in natural language processing literature, is an exponential of the cross-entropy, 
\begin{equation}
\mathcal{P} = 2^{H(Q, P)}
\end{equation}
Of course, the true probability distribution $Q(a)$ is not known, but it can be shown  \cite{Schutze,Martin} that for a corpus of size $M$ obtained from a stationary and ergodic Markov chain, the cross-entropy is given by the limit,
\begin{equation}
H(Q, P) = \lim_{M\rightarrow\infty} {1\over M} -\sum_a \log_{2} P(a)
\end{equation}
In other words, the cross-entropy of the corpus, as the corpus size approaches infinity becomes the cross-entropy of the probability distribution $Q(a)$ from which the corpus was generated. The previous formula does not require knowledge of $Q(a)$, and can then be used to give an estimate of the cross-entropy for a large, but finite, corpus.  The relation has obvious generalisations to joint probability distributions $P(ab)$, $P(abc) \ldots $, of bigrams, trigrams and higher $n$-grams. We denote the $n$-gram cross-entropy by $H_n(Q, P)$. 

Here, we measure the cross-entropy against a held out corpus. We see that the perplexity is reduced considerably when bigram correlations are taken into account. We have also evaluated the perplexity for trigram and quadrigram models, and this shows a monotonic reduction in perplexity as shown in Table ~\ref{tab:Perplexity}. This implies that correlations beyond the preceding sign are important, though the most important correlations comes from the preceeding sign. The perplexity of the bigram model is 26.69 which is significantly lower than that of unigram model which equals 68.82. As discussed in the beginning of this section, this motivates our choice of retaining  only bigram correlations for the present study.

The bigram model can now be used to generate texts according to the Markov chain defined by the unigram and bigram probabilities. In Table ~\ref{tab:GeneratedTexts} we show examples of texts which have been generated by the bigram model. The performance of the model is discussed in Section~\ref{sec:performance}.

Our results in this section establish without doubt that there are significant correlations between signs in the Indus script, and reveal the presence of what can reasonably be called syntax. There are well-defined text beginners and text enders, sign order is important, and many statistically significant pairs of signs occur in the corpus. 

\section{Restoring illegible signs}
\label{sec:filling}

An important practical use of the bigram model is to restore signs which are not legible in the corpus due to damage or other reasons. The basic idea is to use the bigram model to evaluate the probability of a suggested restoration, and chose the restoration with the highest probability. For example, consider the three sign text $\mathcal{S}_3 = s_1s_xs_3$ in which the middle sign $s_x$ is illegible. We use the bigram model to evaluate the probability of the string for different choices of $s_x$ by
\begin{equation}
P({\mathcal S}_3) = P(\texttt{</s>}|s_3)P(s_3|s_x)P(s_x|s_1)P(s_1|\texttt{<s>})
\end{equation}
The most probable sign with which to restore the text is, then, the maximum of this probability over all possible signs $s_x$. Since there are $417$ possible signs, this can be accomplished by a direct enumeration. When the number of illegible signs is more, the space over which the maximisation needs to be done grows rapidly. With $p$ illegible signs, there are $417^p$ values from which to pick a maximum. In such instances, instead of a direct search, a dynamic programming algorithm may be applied. Here, we use the well-known Viterbi algorithm\cite{Schutze} for finding the most probable state sequence for a given observation in a hidden Markov model, suitably modified to make it applicable to a Markov chain, to find most probable sequence of signs. Our results for sign restorations are summarised in Table ~\ref{tab:FillingTable}. We list the original text, a randomly chosen deletion for that text, the most probable restoration, and the next probable restorations obtained using the bigram model. We see that in all cases, the bigram model is successful in reproducing the deleted sign. This gives us confidence that the bigram model can be used to suggest restorations of illegible signs in various corpora. Table ~\ref{tab:illegible} gives few examples of how the model can be used for restoration of doubtfully read signs in the texts of M77 corpus.

\begin{figure}[]
\center
	\includegraphics[width=8cm]{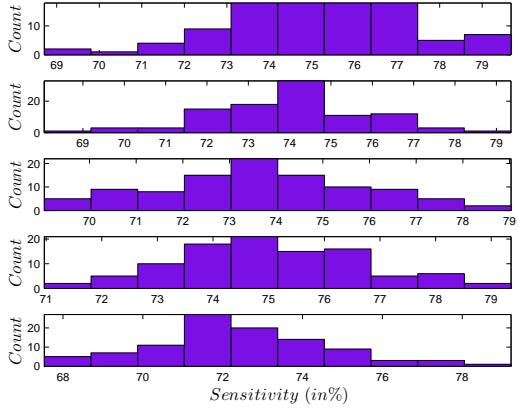}
 	\caption{\label{fig:wbCrossVal90_1}Sensitivity of the bigram model taking all signs under 90\% area of the cumulative probability curve as true positives is 74\%. The five plots are for five different sets of test and training sets of EBUDS as given in Table~\ref{tab:Sensitivity}}
\end{figure}
 
\begin{table}[]
\center
\caption{\label{tab:Sensitivity}Mean Sensitivity (in \%) with Standard Deviation of the model predicted from each of the five test sets $P1$, $P2$, $P3$, $P4$ and $P5$}
\begin{tabular}{cccc}
\hline
Test Set& Train Set & Mean Sensitivity & STDEV\\
\hline
$P1$&\{$P2$, $P3$, $P4$, $P5$\}&$75$&$2$\\
$P2$&\{$P1$, $P3$, $P4$, $P5$\}&$74$&$2$\\
$P3$&\{$P1$, $P2$, $P4$, $P5$\}&$74$&$2$\\
$P4$&\{$P1$, $P2$, $P3$, $P5$\}&$75$&$2$\\
$P5$&\{$P1$, $P2$, $P3$, $P4$\}&$72$&$2$\\
\hline
&Average&74&2\\
\hline
\end{tabular}
\end{table}

We can also use the model to find most probable texts of various lengths. In Table ~\ref{tab:mostProbable} we reproduce the most probable texts of length $4, 5$ and $6$ as predicted by the bigram model. It is quite remarkable that there are exact instances of the most probable texts of length $4, 5$ and slight variant of most probable text of length $6$ in the corpus. 

\begin{table*}[tp]
\caption{\label{tab:mostProbable}The most probable texts of length $4$, $5$ and $6$ predicted by the model. Remarkably, exact instances of the predicted texts are present in the corpus for the $4$-sign and $5$-sign texts. For the $6$-sign text, the same sequence, but with 2 insertions, is found in the corpus.}
\centering
\begin{tabular}{lrrrr}
\hline\\
Text length & Text with blank positions & Most probable predicted text & Text No. & Closest matching text from M77\\\\
\hline\\
\newline
4&\begin{minipage}{0.3in}\texttt{</s>}\end{minipage}\begin{minipage}{0.3in}\hspace{.75cm}\includegraphics[scale=.5]{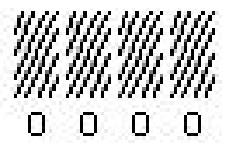}\end{minipage}\begin{minipage}{0.4in}\hspace{4mm}\texttt{<s>}\end{minipage}&\begin{minipage}{2in}\hspace{3.2cm}\includegraphics[scale=.5]{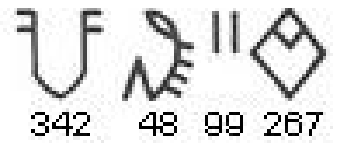}\end{minipage}&1232&\begin{minipage}{2in}\hspace{3.25cm}\includegraphics[scale=.5]{Text4}\end{minipage}\\\\
\newline
&&&2580&\begin{minipage}{2in}\hspace{3.25cm}\includegraphics[scale=.5]{Text4}\end{minipage}\\\\
\newline
5&\begin{minipage}{0.3in}\hspace{3mm}\texttt{</s>}\end{minipage}\begin{minipage}{0.3in}\hspace{0.75mm}\includegraphics[scale=.5]{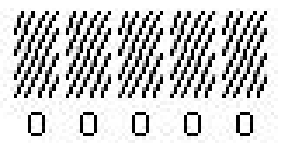}\end{minipage}\begin{minipage}{0.5in}\hspace{7mm}\texttt{<s>}\end{minipage}&\begin{minipage}{2in}\hspace{3cm}\includegraphics[scale=.5]{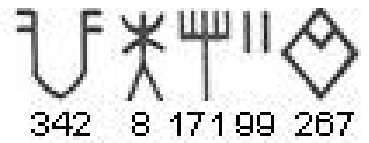}\end{minipage}& 2476 &\begin{minipage}{2in}\hspace{3cm}\includegraphics[scale=.5]{Text5}\end{minipage}\\\\
\newline
6&\begin{minipage}{0.3in}\hspace{5mm}\texttt{</s>}\end{minipage}\begin{minipage}{0.4in}\hspace{3mm}\includegraphics[scale=.5]{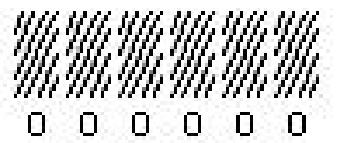}\end{minipage}\begin{minipage}{0.5in}\hspace{7mm}\texttt{<s>}\end{minipage}&\begin{minipage}{2in}\hspace{2.5cm}\includegraphics[scale=.5]{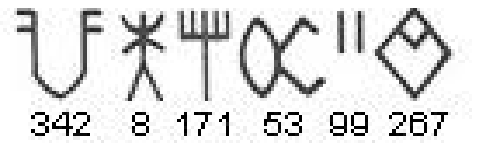}\end{minipage} &1322&\begin{minipage}{2in}\hspace{1.5cm}\includegraphics[scale=.5]{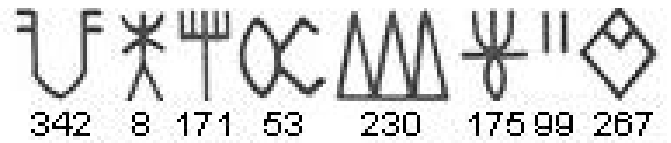}\end{minipage}\\\\
\hline
\end{tabular}
\label{tab:gt}
\end{table*}

\begin{table*}[bp]
\caption{\label{tab:results}Major conclusions}
\begin{tabular}{cllcl}
\hline\\
Sl. No & Test/Measure & Results & Fig. / Table No.& Conclusions\\\\
\hline\\
1. & Zipf-Mandelbrot law  & Best fit for a= 15.4, b = 2.6, &Fig.~\ref{fig:EBUDSZipfsPlot1} & Small number of signs account for \\
  &                       & c = 44.5 (95\% Confidence interval)                            & &bulk of the data while a large number\\
  & & & &  of signs contribute to a long tail.\\\\
2. & Cumulative frequency  & 69 signs: 80 \% of EBUDS &Fig.~\ref{fig:EBUDSCumGraph1} & Indicates asymmetry in usage of 417 \\
  & distribution          & 23 signs: 80 \% of Text Enders & &distinct signs. Suggests directionality \\
  &                       & 82 signs: 80 \% of Text Beginners & & and structure in writing.\\\\
3.& Bigram probability    & Conditional probability matrix for &Fig.~\ref{fig:CondProbMatrices}  & Indicates presence of significant \\      
&                       & EBUDS is strikingly different from &                                 & correlations between signs.\\
&                        & the matrix assuming no correlations.&                                  &\\\\
4. & Conditional probabilities       & Restricted number of signs follow&Fig.~\ref{fig:wbBegEnderAnalsisPlots12}&Indicates presence of signs having \\
   & of text beginners and        & frequent text beginners whereas              &&similar syntactic functions.\\
  & text enders                       & large number of signs precede               && \\
  &                       & frequent text enders.                        && \\\\
5. &Log-likelihood& Significant sign pairs extracted. & Table~\ref{tab:Significance}& The most significant sign pairs are not \\
    &  significance test &                         &                             &always the most frequent ones. \\\\
6. & Entropy               & Random: 8.73; EBUDS: 6.68  & Table~\ref{tab:Entropy}      &Indicates presence of correlations. \\\\
7. & Mutual information    & Random: 0; EBUDS: 2.24     &Table~\ref{tab:Entropy}       &Indicates flexibility in sign usage. \\\\
8. & Perplexity           & Monotonic reduction as n  &Table~\ref{tab:Perplexity}&Indicates presence of long range \\
& &increases from 1 to 5. &&correlations, see also \cite{Yadav1, Yadav2}.\\\\
9. & Sign Restoration     & Restoration of missing and                            & Tables~\ref{tab:FillingTable}, ~\ref{tab:illegible}&Model can restore illegible signs, \\
&&illegible signs.&&according to probability.\\\\
 10. & Cross validation & Sensitivity of the model = 74 \%&Table~\ref{tab:Sensitivity}, Fig.~\ref{fig:wbCrossVal90_1}&Model  can predict signs with \\
 &&&&74\% accuracy.\\
 \hline
\end{tabular}
\end{table*}

\section{Model performance evaluation: Cross Validation}
\label{sec:performance}
We have already discussed {\it perplexity} in section~\ref{sec:bigram}. Here we discuss another method to evaluate the performance of the model called  {\it cross validation}. In cross validation, the corpus is divided into a training set, from which the probabilities are estimated, and a test set, on which the model is evaluated. The measure of goodness of the model calculated is sensitivity which is defined as 
\begin{equation}
\mbox{Sensitivity} = TP/(TP + FN)
\end{equation}
where $TP$ is the count of true positives and $FN$ is the count of false negatives. The ratio of training to test set size used is $80$ : $20$. We divide EBUDS into $5$ equal parts i.e. $P1$, $P2$, $P3$, $P4$ and $P5$ each being one-fifth of the corpus. We start with first part, selecting that to be the test set and concatenate the remaining parts to form the first training set. We use the training set to learn the parameters of the model and use the test set to evaluate the goodness of the model. We drop out signs randomly from the test set and ask the model to fill in the dropped signs. We then count the number of true positives, that is, the number of times the predicted signs matches with the signs under 90\% area of the probability curve; otherwise, they are considered false negatives. This is repeated about 100 times with the first test set and training set and we plot the histogram of the sensitivity for these $100$ runs. 

The cross validation test described above is repeated by taking each of the five equal parts of EBUDS as test set and concatenating remaining parts as training sets. The results are shown in Table~\ref{tab:Sensitivity} and Fig.~\ref{fig:wbCrossVal90_1}. As can be seen from the plots the sensitivity of the model considering all signs under 90\% area of the cumulative probability curve as true positives is 74\%.
 
\section{Summary and conclusions}

We conclude that a bigram model, equivalent to a first-order Markov chain, can be used to capture the correlations of signs in the Indus texts and gives important insights into the structure of Indus writing. Our study has revealed that the script carries syntax, with well-defined text beginners and text enders, directionality and sign order, and strong correlations between signs. Information theoretic measures indicate that the correlations are not completely rigid. This indicates that the script can certainly be considered as a formal language, but it remains to be seen if these features imply an underlying natural language. The tests from which we draw this conclusion are summarised in Table~\ref{tab:results}. The bigram model can be used to suggest probable restorations of illegible signs from a corpus. 

To the best of our knowledge, our work represents the first use of the methods of n-gram analysis to an undeciphered scipt. We believe stochastic methods hold considerable promise in elucidating syntax in undeciphered scripts. We will present further results on the syntax of Indus script, based on stochastic methods, in our forthcoming work.

\section*{Acknowledgements}

We wish to acknowledge the research grant of the Jamsetji Tata Trust that enabled us to do this work. We wish to acknowledge the generous assistance of Indus Research Centre of the Roja Muthiah Research Library. We sincerely thank Prof. P. P. Divakaran for enabling a very fruitful collaboration between TIFR and IMSc. R. Adhikari gratefully acknowledges Pinaki Mitra for first arousing his interest in probability theory and the study of language. Rajesh Rao would like to acknowledge the support of the National Science Foundation (NSF) and the Packard Foundation. We are grateful to Harappa.com for their kind permission to use the picture of an Indus seal in the paper.
\eject
\bibliographystyle{IEEETran_ASP}
\bibliography{apssampIndus}

\end{document}